\documentclass[11pt]{article}

\usepackage[truedimen,margin=25truemm]{geometry}

\usepackage{graphicx} 
\usepackage{epsfig} 
\usepackage{subfigure} 

\usepackage{lscape}

\usepackage{amsmath}
\usepackage{amsthm}
\usepackage{amssymb}
\usepackage{ascmac}
\usepackage{algorithm}
\usepackage{algpseudocode}

\usepackage{color}

\usepackage{ulem}

\usepackage{booktabs}

\usepackage{hyperref}

\usepackage{graphicx} 
\usepackage{epsfig}
\usepackage{subfigure}

\usepackage{amssymb}
\usepackage{amsmath}
\usepackage{amsthm}
\usepackage{mathtools}
\usepackage[raggedright]{titlesec}

\def\vec#1{\mbox{\boldmath $#1$}}
\def\mat#1{\mbox{\bf #1}}

\newcommand{\argmin}{\mathop{\rm arg~min}\limits}

\title{Wasserstein $k$-means with sparse simplex projection}

\author{Takumi Fukunaga \thanks{Department of Communications and Computer Engineering, School of Fundamental Science and Engineering, WASEDA University, 3-4-1 Okubo, Shinjuku-ku, Tokyo 169-8555, Japan (e-mail: f\_takumi1997@suou.waseda.jp)} \and Hiroyuki Kasai \thanks{Department of Communications and Computer Engineering, School of Fundamental Science and Engineering, WASEDA University, 3-4-1 Okubo, Shinjuku-ku, Tokyo 169-8555, Japan (e-mail: hiroyuki.kasai@waseda.jp)}}

\begin{document}

\maketitle

\begin{abstract}
This paper presents a proposal of a faster Wasserstein $k$-means algorithm for histogram data by reducing Wasserstein distance computations and exploiting sparse simplex projection. We shrink data samples, centroids, and the ground cost matrix, which leads to considerable reduction of the computations used to solve optimal transport problems without loss of clustering quality. Furthermore, we dynamically reduced the computational complexity by removing lower-valued data samples and harnessing sparse simplex projection while keeping the degradation of clustering quality lower. We designate this proposed algorithm as sparse simplex projection based Wasserstein $k$-means, or SSPW $k$-means. Numerical evaluations conducted with comparison to results obtained using Wasserstein $k$-means algorithm demonstrate the effectiveness of the proposed SSPW $k$-means for real-world datasets \footnote{This paper has been accepted in ICPR2020 \cite{fukunaga_icpr_2020}.}.
\end{abstract}

\section{Introduction}
The $k$-means clustering algorithm is a simple yet powerful algorithm, which has been widely used in various application areas including, for example, computer vision, statistic, robotics, bioinformatics, machine learning, signal processing and medical image engineering. The standard Euclidean $k$-means algorithm proposed by Lloyd \cite{Lloyd_IEEEST_1982_s} minimizes the total squared distances between all data samples and their assigned cluster {\it centroids}. This algorithm consists of two iterative steps, the assignment step and the update step, when clustering $q$ data samples into $k$ clusters. Because the assignment step requires calculation cost $\mathcal{O}(qk)$, this step becomes prohibitively expensive when $q$ is larger. For this issue, many authors have proposed efficient approaches \cite{Alsabti_IPPS_1998, Pelleg_KDD_1999, Kanungo_PAMI_2002,Elkan_ICML_2003, Hamerly_SDM_2010, Drake_OPT_2012, Ding_ICML_2015,Bottesch_ICML_2016}. 
It should be also noted that $k$-means algorithm is equivalent to non-negative matrix factorization (NMF) \cite{Chris_SDM_2005}, and NMF has been applied to many fields \cite{Innami_CMA_2012,Kasai_IEEEICASSP_2018,Kasai_EUSIPCO_2018,Hashimoto_ICASSP_2020}.
Regarding another direction of the research in the $k$-means clustering algorithm, some efforts have achieved improvement of the clustering quality. Some of them address the point that the $k$-means clustering algorithm ignores {\it latent} structure of data samples, such as {\it histogram} data \cite{Bottesch_ICML_2016}. One approach in this category achieves better performances than the standard algorithm in terms of the clustering quality by exploiting the {\it Wasserstein} distance \cite{Ye_2017_IEEETSP_s}, a metric between probability distributions of histograms which respects their latent geometry of the space. This is called the {\it Wasserstein $k$-means} algorithm hereinafter. The Wasserstein distance, which originates from {\it optimal transport} theory \cite{Peyre_2019_OTBook}, has many favorable properties \cite{Villani_2008_OTBook}. Therefore, it is applied for a wide variety problems in machine learning \cite{Kasai_ICASSP_2020}, statistical learning and signal processing. Nevertheless, calculation of the Wasserstein distance requires much higher computational costs than that of the standard Euclidean $k$-means algorithm because it calculates an optimization problem every iteration, which makes the Wasserstein $k$-means infeasible for a larger-scale data. 

To this end, this paper presents a proposal of a faster Wasserstein $k$-means algorithm for histogram data by reducing the Wasserstein distance computations exploiting {\it sparse simplex projection}. More concretely, we shrink data samples, centroids, and the ground cost matrix, which drastically reduces computations to solve linear programming optimization problems in optimal transport problems without loss of clustering quality. Furthermore, we dynamically reduce the computational complexity by removing lower-valued histogram elements and harnessing the sparse simplex projection while maintaining degradation of the clustering quality lower. We designate this proposed algorithm as sparse simplex projection based Wasserstein $k$-means, or SSPW $k$-means. Numerical evaluations are then conducted against the standard Wasserstein $k$-means algorithm to demonstrate the effectiveness of the proposed SSPW $k$-means for real-world datasets.

The source code will be made available at \url{https://github.com/hiroyuki-kasai}.

\section{Preliminaries}
This section first presents the notation used in the remainder of this paper. It then introduces the optimal transport problem, followed by explanation of the Wasserstein $k$-means algorithm.

\subsection{Notation}
We denote scalars with lower-case letters $(a, b, \ldots)$, vectors as bold lower-case letters $(\vec{a}, \vec{b}, \ldots)$, and matrices as bold-face capitals $(\mat{A}, \mat{B}, \ldots)$. The $i$-th element of \vec{a} and the element at the $(i,j)$ position of \mat{A} are represented as $\vec{a}_i$ and $\mat{A}_{ij}$, respectively. $\vec{1}_d$ is used for the $d$-dimensional vector of ones. Operator $(\cdot)^T$ stands for the matrix transpose. Operator $[\vec{a}_i]_+$ represents $\mathrm{max}(a,0)$ that outputs $a$ when $a\! \geq\! 0$, and $0$ otherwise. Given a set $\mathcal{S} \subseteq \mathcal{N} = \{1, \ldots, n\}$, the complement $\mathcal{S}^c$ is defined with respect to $\mathcal{N}$, and the cardinality is $|\mathcal{S}|$. The support set of \vec{a} is ${\rm supp}(\vec{a}) = \{i : \vec{a}_i \neq 0\}$. $\vec{a}_{| \mathcal{S}}$ extracts the elements of $\mathcal{S}$ in \vec{a}, of which size is $|\mathcal{S}|$. $\mathbb{R}_{+}^{n \times m}$ represents a nonnegative matrix of size ${n \times m}$. The unit-simplex, simply called {\it simplex}, is denoted by $\Delta_n$, which is the subset of $\mathbb{R}^n$ comprising all nonnegative vectors whose sums is 1. $\lfloor a \rfloor$ represents the floor function, which outputs the greatest integer less than or equal to $a$.

\subsection{Euclidean $k$-means and speed-up methods}
\label{sec:kmeans}

The goal of the standard Euclidean $k$-means algorithm is to partition given $q$ data $\mat{X} = \{\vec{x}_1, \ldots, \vec{x}_q\} \subset \mathbb{R}^d$ into $k$ separated groups, i.e., $k$ {\it clusters}, as $\mathcal{C}= \{C_1, \ldots, C_k\}$ such that the following $k$-means cost function is minimized:
\begin{eqnarray}
\label{Eq:KmeansObj}
	f(\mat{C}, \mat{X}) & = & \sum_{j=1}^k \sum_{\vec{x}_i \in C_j} \| \vec{x}_i - \vec{c}_j \|_2^2,
\end{eqnarray}
where $\mat{C}=\{\vec{c}_1, \ldots, \vec{c}_k\}\subset \mathbb{R}^d$ is a set of $k$ cluster centers, designated as centroids hereinafter. Also, $ \| \cdot \|_2^2$ represents the squared Euclidean distance. Consequently, our aim is to find $k$ cluster centroids $\mat{C}$. It is noteworthy that the problem defined in (\ref{Eq:KmeansObj}) is {\it non-convex}. It is known to be an NP-hard problem \cite{Dasgupta_TR_2007, Mahajan_WALCOM_2009}. Therefore, many efforts have been undertaken to seek a {\it locally} optimized solution of (\ref{Eq:KmeansObj}). Among them, the most popular but simple and powerful algorithm is the one proposed by Lloyd \cite{Lloyd_IEEEST_1982_s}, called Lloyd's $k$-means. 
This algorithm consists of two iterative steps where the first step, called the {\it assignment} step, finds the closest cluster centroid $\vec{c}_j (\in [k])$ for each data sample $\vec{x}_i (\in [q])$. It then updates the cluster centroids $\mat{C}$ at the second step, called the {\it update} step. These two steps continue until the clusters do not change anymore. The algorithm is presented in {\bf Algorithm \ref{LoydAlgorithm}}. 

\begin{algorithm} 
\caption{Lloyd's algorithm for $k$-means \cite{Lloyd_IEEEST_1982_s}}
\label{LoydAlgorithm}
\begin{algorithmic}[1]
\Require{data $\mat{X}\!=\!\{\vec{x}_1, \ldots, \vec{x}_q\} \!\subset\! \mathbb{R}^d$, cluster number $k \in \mathbb{N}$.} 
\State{Initialize randomly centroids $\mat{C}=\{\vec{c}_1, \ldots, \vec{c}_k\}\subset \mathbb{R}^d$.}
\Repeat
\State{Find closest centroids (assignment step):\\
\hspace*{1cm} $s_i= \text{argmin}_{j=1, \ldots, k}\ \|\vec{x}_i-\vec{c}_j\|_2^2, \forall i \in [q].$}
\State{Update centroids (update step):\\
\hspace*{1cm} $\vec{c}_j = {\rm mean}(\vec{x} \in \mat{X} | s_i=j\}),  \forall j \in [k].$}
\Until {cluster centroids stop changing.}
\Ensure{cluster centers $\mat{C}=\{\vec{c}_1, \ldots, \vec{c}_k\}$.}
\end{algorithmic}
\end{algorithm}

Although Lloyd's $k$-means is simple and powerful, the obtained result is not guaranteed to reach a {\it global} optimum because the problem is not convex. For that reason, several efforts have been launched to find good initializations of Lloyd's algorithm \cite{Arthur_SODA_2007, Ostrovsky_FOCS_2006, Bahmani_VLDB_2012}.
Apart from the lack of the global guarantee, Lloyd's algorithm is adversely affected by high computation complexity. The highest complexity comes from the assignment step, where the Euclidean distances among all the data samples and all cluster centroids must be calculated. 
This cost of complexity is $\mathcal{O}(qk)$, which is infeasible for numerous data samples or clusters. It is also problematic for higher dimension $d$. For this particular problem, many authors have proposed efficient approaches. One category of approaches is to generate a hierarchical tree structure of the data samples and to make the assignment step more efficient \cite{Alsabti_IPPS_1998, Pelleg_KDD_1999, Kanungo_PAMI_2002}. However, this also entails high costs to maintain this structure. It becomes prohibitively expensive for larger dimension data samples. Another category is to reduce {\it redundant} calculations of the distance at the assignment step. For example, if the assigned centroid changes closer to a data sample $\vec{x}$ in {\it successive} iterations, then it is readily apparent that all other centroids that did not change cannot move to $\vec{x}$. Consequently, the distance calculations to the corresponding centroids can be omitted. In addition, the triangle inequality is useful to skip the calculation. This category includes, for example, \cite{Elkan_ICML_2003, Hamerly_SDM_2010, Drake_OPT_2012, Ding_ICML_2015,Bottesch_ICML_2016}. However, it remains unclear whether these approaches are applicable to Wasserstein $k$-means described later in {Section \ref{sec:Wkmeans}}.

\subsection{Optimal transport \cite{Peyre_2019_OTBook}}
\label{Sec:OT}
Let $\Omega$ be an arbitrary space, $d(\cdot, \cdot)$ a {\it metric} on that space, and $P(\Omega)$ the set of Borel {\it probability measures} on $\Omega$. For any point $\nu \in \Omega$ is the Dirac unit mass on $\nu$. Here, we consider two families $\vec{\nu}=(\nu_1, \ldots, \nu_m)$ and $\vec{\mu}=(\mu_1, \ldots, \mu_n)$ of point in $\Omega$, where $\vec{a}$ and $\vec{b}$ respectively satisfy $\vec{a}\!=\!(a_1 \ldots, a_m)^T \in \mathbb{R}_+^m$ and $\vec{b}\!=\!(b_1 \ldots, b_n)^T \in \mathbb{R}_+^n$ with $\sum_i^m a_i=\sum_j^n b_j = 1$. The {\it ground cost matrix} $\mat{C}_{\scriptsize \vec{\nu \mu}}$, or simply \mat{C}, represents the distances between elements of \vec{\nu} and \vec{\mu} raised to the power $p$ as 
%
	$\mat{C}_{{\scriptsize{\vec{\nu \mu}}}} \ =\ \mat{C}  := [d(\nu_u, \mu_v)^p]_{uv} \ \ \in \mathbb{R}_+^{m \times n}$.
%
The cost {\it transportation polytope} $\mathcal{U}_{mn}$ of $a \in \Delta_m$ and $b \in \Delta_n$ is a feasible set defined as the set of $m \times n$ nonnegative matrices such that their row and column {\it marginals} are respectively equal to $a$ and $b$. It is formally defined as
\begin{eqnarray*}
	\mathcal{U}_{mn} &:=& 
	\{
	\mat{T} \in \mathbb{R}^{m\times n}_+ : \mat{T} \vec{1}_n = \vec{a}, \mat{T}^T \vec{1}_m = \vec{b}
	\}.
\end{eqnarray*}

Consequently, the {\it optimal transport matrix} $\mat{T}^*$ is defined as a solution of the following {\it optimal transport} problem \cite{Peyre_2019_OTBook}
\begin{eqnarray}
\label{Eq:optimal_transport_problem}
	\mat{T}^* &=& \argmin_{\scriptsize \mat{T}\ \in\ \mathcal{U}_{mn}}\ \langle \mat{T}, \mat{C}\rangle.
\end{eqnarray}

\subsection{Wasserstein distance and Wasserstein $k$-means \cite{Ye_2017_IEEETSP_s}}
\label{sec:Wkmeans}

The Wasserstein $k$-means algorithm replaces the distance metric $\|\vec{x}_i-\vec{c}_j\|_2^2$ in (\ref{Eq:KmeansObj}) with the following the $p$-Wasserstein distance $W_p(\vec{\mu}, \vec{\nu})$, which is formally defined as shown below.

\noindent
{\bf Definition 1}\ ($p$-Wasserstein distance){\bf .}
For $\vec{\mu}$ and $\vec{\nu}$, the Wasserstein distance of order $p$ is defined as 
\begin{eqnarray*}
	W_p(\vec{\mu}, \vec{\nu}) 
	& = & \min_{\scriptsize \mat{T}\ \in\ \mathcal{U}_{mn}}\ \langle \mat{T}, \mat{C}\rangle 
 	 =  \langle \mat{T}^*, \mat{C} \rangle.
\end{eqnarray*}

In addition, the calculation in Step 2 in {\bf Algorithm \ref{LoydAlgorithm}}, i.e., the update step, is replaced with the Wasserstein barycenter calculation, which is defined as presented below.

\noindent
{\bf Definition 2}\ (Wasserstein barycenter \cite{Benamou_2015_SIAMJSC}){\bf .}
A Wasserstein barycenter of $q$ measures $\{\vec{\nu}_1, \ldots, \vec{\nu}_q\} \in \mathcal{P} \subset P(\Omega) $ is a minimizer of $g(\vec{\mu})$ over $\mathcal{P}$ as
\begin{eqnarray*}
	g(\vec{\mu}) &:=& \frac{1}{n} \sum_{i=1}^q W_p(\vec{\mu}, \vec{\nu}_i).
\end{eqnarray*}

Addressing that $m=n$ holds in this case, we assume $\{\mat{T}, \mat{C}\} \subset \mathbb{R}^{n\times n}_+$ hereinafter. Then, the overall algorithm is summarized in {\bf Algorithm \ref{Alg:Wkmeans}}. 

\begin{algorithm}[t]
\caption{Wasserstein $k$-means}
\label{Alg:Wkmeans}
\begin{algorithmic}[1]
\Require{data $\{\vec{\nu}_1, \ldots, \vec{\nu}_q\}$, cluster number $k \in \mathbb{N}$, ground cost matrix $\mat{C} \in \mathbb{R}^{n \times n}$.}
\State{Initialize centroids $\{{\vec{c}}_1, \ldots, {\vec{c}}_k\}$.}
\Repeat
\State{Find closest centroids (assignment step):\\
\hspace*{1cm} $s_i = \text{argmin}_{j=1, \ldots, k}\ W_p({\vec{\nu}}_i, {\vec{c}}_j), \forall i \in [q].$}
\State{Update centroids (update step):\\
\hspace*{1cm} $\vec{c}_j = \text{barycenter}(\{\vec{\nu}|s_i=j\}), \forall j \in [k].$}
\Until {cluster centroids stop changing.}
\Ensure{cluster centers $\{\vec{c}_1, \ldots, \vec{c}_k\}$.}
\end{algorithmic}
\end{algorithm}

The computation of the Wasserstein barycenters has recently gained increasing attention in the literature because it has a wide range of applications. Many fast algorithms have been proposed, and they include, for example, \cite{Cuturi_ICML_2014,Bonneel_JMIV_2015,Anderes_MMOR_2016,Ye_2017_IEEETSP_s,Claici_ICML_2018,Puccetti_JMA_2020}. Although they are effective at the update step in the Wasserstein $k$-means algorithm, they do not directly tackle the computational issue at the assignment step that this paper particularly addresses. \cite{Staib_CI_2017} proposes a mini-batch algorithm for the Wasserstein $k$-means. However, it is not clear whether it gives better performances than the original Wasserstein $k$-means algorithm due to the lack of comprehensive experiments. Regarding another direction, the direct calculation of a Wasserstein barycenter of sparse support may give better performances on the problem of interest in this paper. However, as suggested and proved in \cite{Borgwardt_arXiv_2019}, the finding such a sparse barycenter is a hard problem, even in dimension $2$ and for only $3$ measures.

\section{Sparse simplex projection-based Wasserstein $k$-means: SSPW $k$-means}
After this section describes the motivation of the proposed method, it presents elaboration of the algorithm of the proposed sparse simplex projection based Wasserstein $k$-means.

\subsection{Motivation}

The optimal transport problem in (\ref{Eq:optimal_transport_problem}) described in {Section \ref{Sec:OT}} is reformulated as 
\begin{align}
\label{Eq:optimal_transport_LP_problem}
	\mat{T}^* &= \argmin_{\scriptsize \mat{T}\ \in\ \mathcal{U}_{mn}}\ [\mat{T}]_{uv} \| \nu_u - \mu_v \|_p^p,\notag\\
	{\rm s.t.} & \quad  \sum_{u=1}^m [\mat{T}]_{uv} = b_v, v \in [n],\sum_{j=1}^n [\mat{T}]_{uv} = a_u, u \in [m]. 
\end{align}

Consequently, this problem is reduced to the {\it linear programming} (LP) problem with $(m+n)$ {\it linear} equality constraints. The LP problem is an optimization problem with a linear objective function and linear equality and inequality constraints. The problem is solvable using general LP problem solvers such as the simplex method, the interior-point method and those variants. The computational complexity of those methods, however, scales at best {\it cubically} in the size of the input data. Here, if the measures have $ n (=m)$ bits, then the number of unknowns $[\mat{T}]_{uv}$ is $n^2$, and the computational complexities of the solvers are at best $\mathcal{O}(n^3 \log n)$ \cite{Cuturi_2013_NIPS,RUBNER_IJCV_2000}. To alleviate this difficulty, the entropy-regularized algorithm, known as {\it Sinkhorn algorithm}, is proposed under the entropy-regularized Wasserstein distance. This algorithm regularizes the Wasserstein metric by the entropy of the transport plan and enables much faster numerical schemes with complexity $\mathcal{O}(n^2)$ or $\mathcal{O}(n \log n)$. It is nevertheless difficult to obtain a higher-accuracy solution. In addition, it can be unstable if the regularization parameter is reduced to a smaller value to improve the accuracy. 

Subsequently, this paper continues to describe the original LP problem in (\ref{Eq:optimal_transport_LP_problem}). We attempt to reduce the size of the measures. More specifically, we adopt the following approaches:
(i) we make the input data samples $\vec{\nu}$ and the centroids $\vec{c}$ sparser than the original one by projecting them onto sparser simplex, and 
(ii) we shrink the projected data sample $\vec{\nu}$ and centroid $\vec{c}$ and the corresponding ground cost matrix $\mat{C}_{\vec{\nu c}}$ by removing their zero elements.
Noteworthy points are that the first approach enables to speed-up the computation while maintaining degradation of the clustering quality as small as possible. The second approach yields no degradation.

\begin{algorithm}[t]
\caption{Proposed sparse simplex projection Wasserstein $k$-means (SSPW $k$-means)}
\label{Alg:SSPWkmeans}
\begin{algorithmic}[1]
\Require{data $\{\vec{\nu}_1,\! \ldots,\! \vec{\nu}_q\}$, cluster number $k\! \in\! \mathbb{N}$, ground cost matrix $\mat{C} \in \mathbb{R}^{n \times n}$, maximum number $T_{\rm max}$, $\gamma_{\rm min}$.}
\State{Initialize centroids $\{\tilde{\vec{c}}_1, \ldots,\tilde{\vec{c}}_k\}$, set $t=1$.}
\Repeat
\State{Update sparsity ratio $\gamma(t)$.}
\State{Project $\vec{\nu}_i$ to $\hat{\vec{\nu}}_i$ on sparse simplex $\Delta_p$:\\
\hspace*{1cm} $\hat{\vec{\nu}}_i = \text{Proj}^{\gamma(t)}(\vec{\nu}_i)\  \forall i \in [q]$}.
\State{Shrink $\hat{\vec{\nu}}_i$ into $\tilde{\vec{\nu}}_i$: $\tilde{\vec{\nu}}_i =\text{shrink}(\hat{\vec{\nu}}_i)$.}
\State{Project $\vec{c}_j$ into $\hat{\vec{c}}_j$ on sparse simplex $\Delta_p$: \\
\hspace*{1cm} $\hat{\vec{c}}_j = \text{Proj}^{\gamma}(\vec{c}_j)\ \forall j \in [k]$.}
\State{Shrink $\hat{\vec{c}}_j$ into $\tilde{\vec{c}}_j$: $\tilde{\vec{c}}_j =\text{shrink}(\hat{\vec{c}}_j)$.}
\State{Shrink ground cost matrix $\mat{C}$ into $\tilde{\mat{C}}$: $\tilde{\mat{C}}=\text{Shrink}(\mat{C})$}
\State{Find closest centroids (assignment step):\\
\hspace*{1cm} $s_i = \text{argmin}_{j=1, \ldots, k}\ W_p(\tilde{\vec{\nu}}_i, \tilde{\vec{c}}_j), \forall i \in [q].$}
\State{Update centroids (update step):\\
\hspace*{1cm} $\vec{c}_j = \text{barycenter}(\{\vec{\nu}|s_i=j\}), \forall j \in [k].$}
\Until {cluster centroids stop changing.}
State{Update the iteration number $t$ as $t=t+1$.}
\Ensure{cluster centers $\{\vec{c}_1, \ldots, \vec{c}_k\}$.}
\end{algorithmic}
\end{algorithm}

\subsection{Sparse simplex projection}
\label{Sec:SSP}
We first consider the sparse simplex projection $\text{Proj}^{\gamma(t)}(\cdot)$, where $\gamma(t) \in (0,1]$ is the control parameter of the sparsity, which is described in {Section \ref{sec:gamma}}. $\text{Proj}^{\gamma(t)}(\cdot)$ projects the $i$-th data sample $\vec{\nu}_i$ and the $j$-th centroid $\vec{c}_j$, respectively, into sparse $\hat{\vec{\nu}}_{i}$ and $\hat{\vec{c}}_j$ on $\Delta_n$. For this purpose, we exploit the projection method proposed in \cite{Kyrillidis_ICML_2013}, which is called GSHP. 

More concretely, denoting the original $\vec{\nu}_i$ or $\vec{c}_j$ as $\vec{\beta} \in \Delta_n$, and the projected $\hat{\vec{\nu}}_{i}$ or $\hat{\vec{c}}_j$ as $\hat{\vec{\beta}}\in \Delta_n$, then $\hat{\vec{\beta}}$ is generated as
\begin{eqnarray*}
\label{Eq:SparseProjection}
\hat{\vec{\beta}} = \text{Proj}^{\gamma(t)}(\vec{\beta}) 
= \left\{
\begin{array}{lrl}
\hat{\vec{\beta}}_{| \mathcal{S}^{\star}} &\!\!=\!\! & \mathcal{P}_{\Delta_{\kappa}}(\vec{\beta}_{|\mathcal{S}^{\star}})\\
\hat{\vec{\beta}}_{| (\mathcal{S}^{\star})^c} &\!\! =\!\! & 0,
\end{array}
\right.
\end{eqnarray*}
where $\kappa=\lfloor n\cdot \gamma(t)\rfloor$, and where $\mathcal{S}^{\star}$ is defined as
	$\mathcal{S}^{\star}  = {\rm supp}(\mathcal{P}_{\kappa}(\vec{\beta}))$,
where $\mathcal{P}_{\kappa}(\vec{\beta})$ is the operator that keeps the $\kappa$-largest elements of \vec{\beta} and sets the rest to zero. Additionally, the $v (\in [n])$-th element of $\mathcal{P}_{\Delta_{\kappa}}(\vec{\beta}_{|\mathcal{S}^{\star}})$ is defined as
\begin{eqnarray*}
\label{Eq:Plambda}
	(\mathcal{P}_{\Delta_{\kappa}}(\vec{\beta}_{|\mathcal{S}^{\star}}))_v &=&  [(\vec{\beta}_{| \mathcal{S}^{\star}})_v + \tau]_+, 
\end{eqnarray*}
where $\tau$ is defined as
\begin{eqnarray*}
	\tau &:=& \frac{1}{\kappa}\left( 1 + \sum^{|\mathcal{S}^{\star}|} \vec{\beta}_{| \mathcal{S}^{\star}} \right).
\end{eqnarray*}	
	
This computational complexity is $ \mathcal{O}(n \min(\kappa, \log(n)))$. 

\begin{center}
\begin{table*}[t]
\caption{Averaged clustering performance results on COIL-100 dataset dataset (5 runs). The best result in each $\gamma_{\rm min}$ is shown in bold. The best in all settings is shown in bold with underline.}

\begin{center}
{
\label{tabl:COIL100_dataset}
\begin{tabular}{c|c|c|c||c|c|c|c|c|c|c|c}
\hline
method & shrink& \multicolumn{2}{|c||}{projection} &  \multicolumn{2}{|c|}{Purity} & \multicolumn{2}{|c|}{NMI} 
& \multicolumn{2}{|c|}{Accurarcy}& \multicolumn{2}{|c}{Time } \\
\cline{3-4}
 &oper. & $\tilde{\vec{\nu}}_i$ & $\tilde{\vec{c}}_j$ & \multicolumn{2}{|c|}{[$\times 10^{2}$]} & \multicolumn{2}{|c|}{[$\times 10^{2}$]} & \multicolumn{2}{|c|}{[$\times 10^{2}$]}& \multicolumn{2}{|c}{[$\times 10^{2}$sec]}  \\
\hline
\hline
$k$-means & & & & \multicolumn{2}{|c|}{{38.0}} & \multicolumn{2}{|c|}{{41.2}} & \multicolumn{2}{|c|}{{35.4}}& \multicolumn{2}{|c}{{$1.49 \times 10^{-4}$}}\\
\hline
\hline
baseline & & & & \multicolumn{2}{|c|}{59.2} & \multicolumn{2}{|c|}{65.5} & \multicolumn{2}{|c|}{58.2}& \multicolumn{2}{|c}{6.51}\\
\cline{1-4} 
\cline{11-12} 
 shrink & $\checkmark$ && & \multicolumn{2}{|c|}{} & \multicolumn{2}{|c|}{} & \multicolumn{2}{|c|}{} & \multicolumn{2}{|c}{5.01} \\
\hline
\hline
 --& --  &\multicolumn{2}{|c||}{$\gamma_{\rm min}$} & 0.7 & 0.8 & 0.7 & 0.8 &0.7 & 0.8 &0.7 & 0.8 \\
\hline
\hline
 & $\checkmark$& & $\checkmark$&
56.5&57.7&
65.2&64.4&
55.0&55.9&
2.50&3.28\\
\cline{2-12}
FIX & $\checkmark$& $\checkmark$&&
61.3&60.9&
66.0&66.6&
59.1&59.7&
2.73&3.60\\
\cline{2-12}
 & $\checkmark$& $\checkmark$& $\checkmark$ &
60.7&57.3&
66.8&65.5&
59.0&55.8&
\underline{\bf 1.62}&{\bf 2.40}\\
\hline
 & $\checkmark$& & $\checkmark$&
59.5&59.7&
65.0&65.6&
58.4&58.6&
2.62&3.43\\
\cline{2-12}
DEC & $\checkmark$& $\checkmark$& &
60.7&59.1&
66.7&65.5&
59.7&58.0&
4.15&4.00\\
\cline{2-12}
 & $\checkmark$& $\checkmark$ & $\checkmark$&
60.0&59.5&
66.0&65.4&
59.0&58.5&
2.83&3.45\\
\hline
 & $\checkmark$& & $\checkmark$&
58.4&58.5&
65.0&65.5&
57.0&57.3&
6.20&6.65\\
\cline{2-12}
INC & $\checkmark$& $\checkmark$& &
\underline{\bf 63.5}&{\bf 61.2}&
\underline{\bf 68.1}&{\bf 66.8}&
\underline{\bf 62.5}&{\bf 59.9}&
6.27&6.76\\
\cline{2-12}
 & $\checkmark$& $\checkmark$ & $\checkmark$&
58.3&57.2&
65.4&64.5&
56.5&55.9&
5.12&5.82\\
\hline
\end{tabular}
}
\end{center}
\end{table*}
\end{center}
\begin{figure*}[t]
\begin{center}
\begin{minipage}[b]{.48\linewidth}
 \centering
\includegraphics[width=1\linewidth]{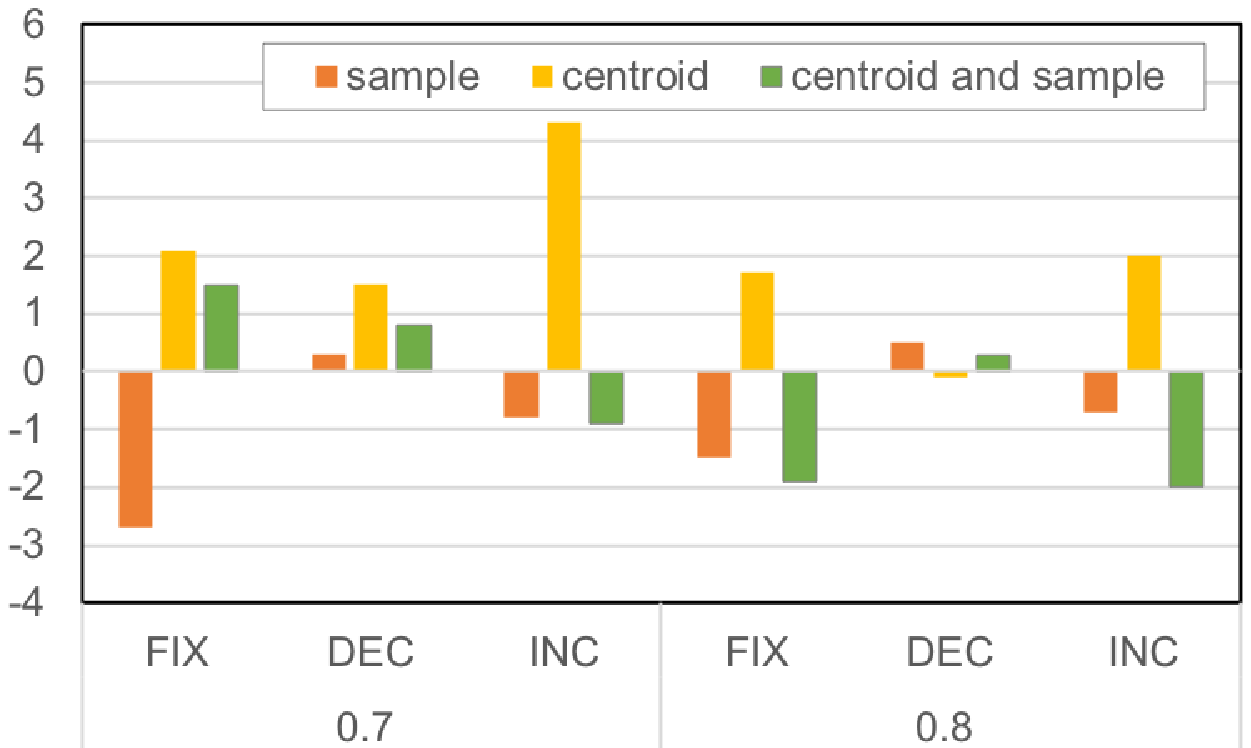}

 {\small (a) Purity}\medskip
\end{minipage}
\begin{minipage}[b]{.48\linewidth}
 \centering
\includegraphics[width=1\linewidth]{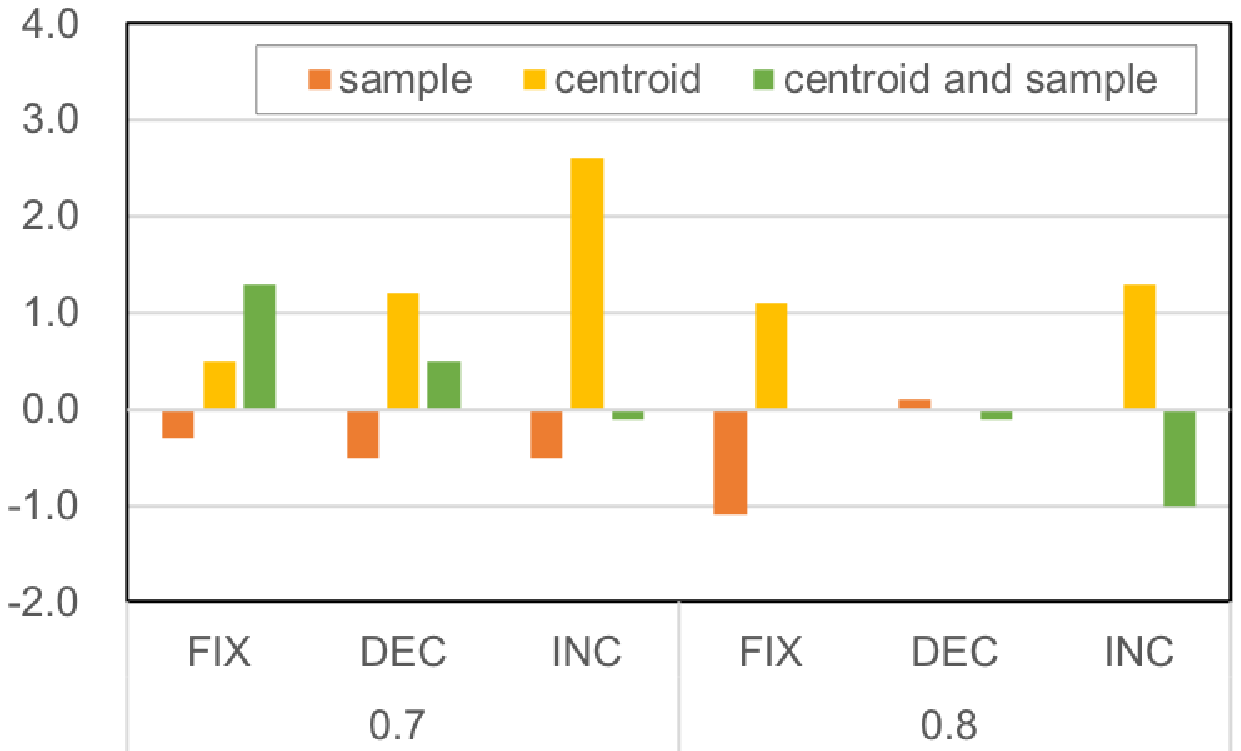}

{\small (b) NMI}\medskip
\end{minipage}

\begin{minipage}[b]{.48\linewidth}
 \centering
\includegraphics[width=1\linewidth]{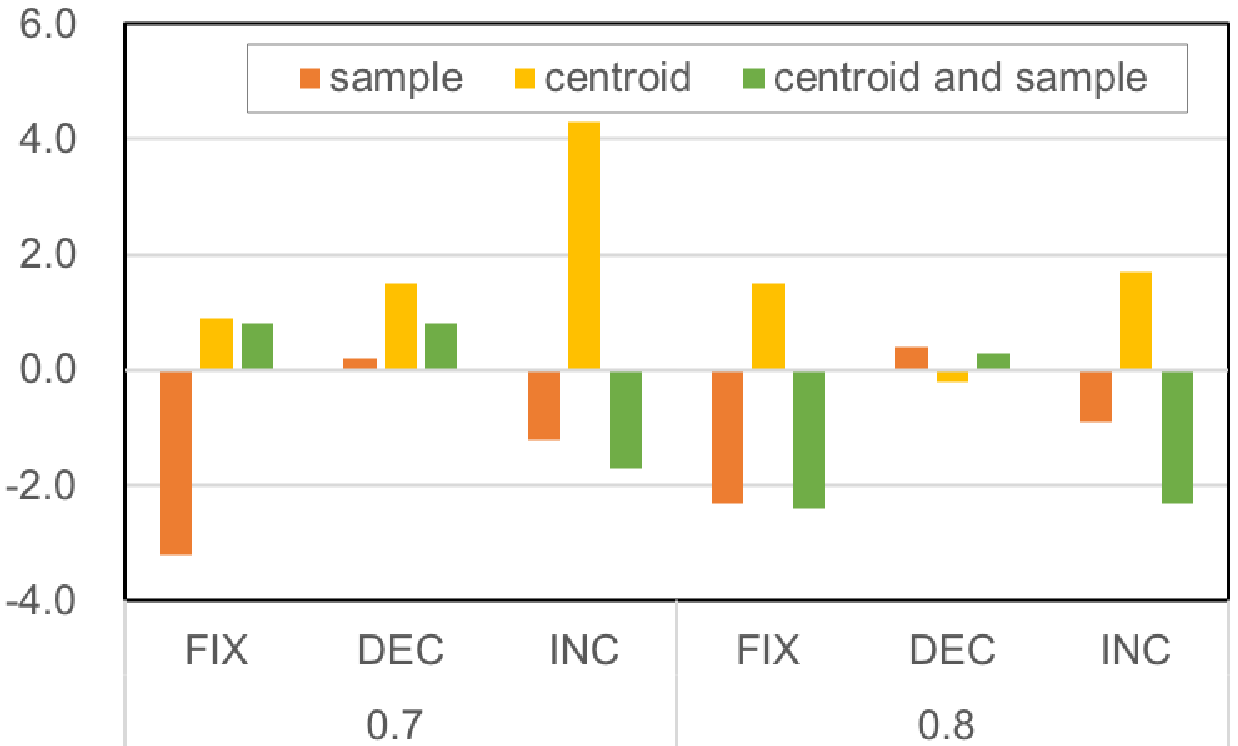}

 {\small (c) Accuracy}\medskip
\end{minipage}
\begin{minipage}[b]{.48\linewidth}
 \centering
\includegraphics[width=1\linewidth]{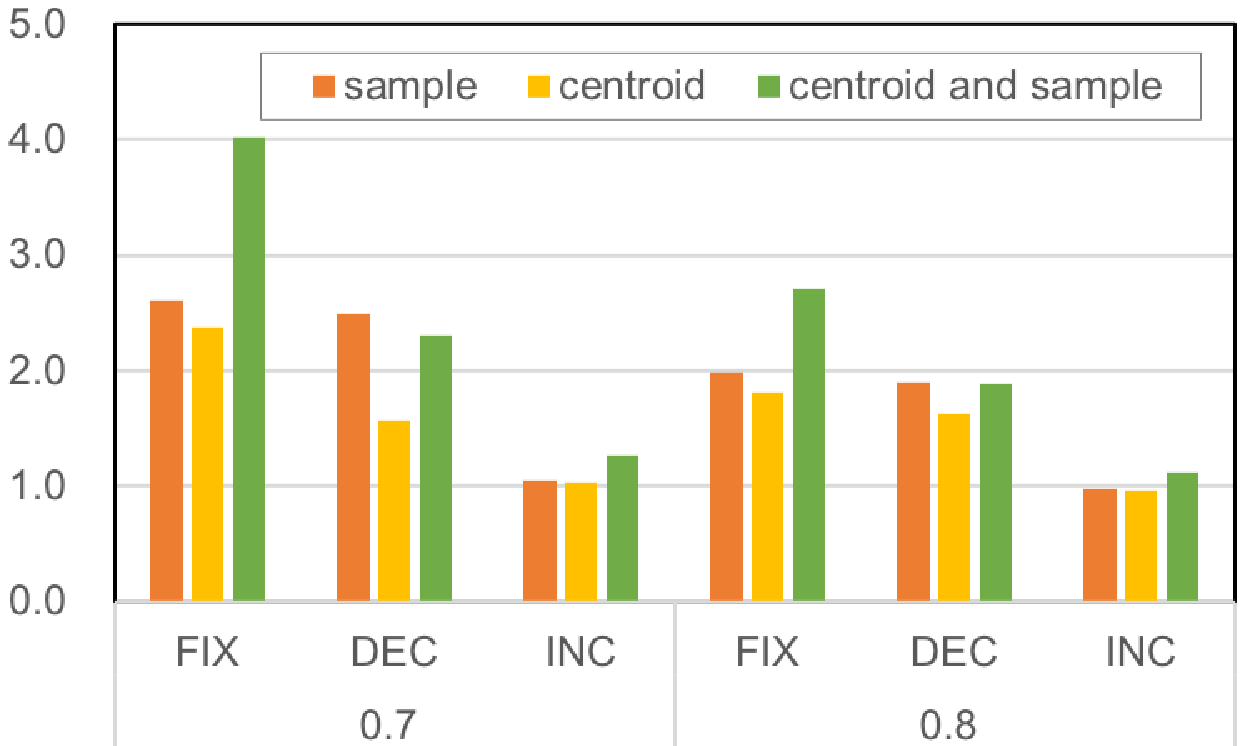}

{\small (d) Computation time}\medskip
\end{minipage}
\end{center}

\caption{Performance results of 1-D histogram data on the COIL100 dataset.}
\label{fig:COIL100_clustering_results}
\end{figure*}

\subsection{Shrinking operations according to zero elements}

The sparse simplex projection in the previous section increases the number of zeros in the centroid and data samples. We reduce the computational complexity by harnessing this sparse structure. The {\it vector shrinking} operator, denoted as $\text{shrink}(\cdot)$, removes the zero elements from the projected sample $\hat{\vec{\nu}}_{i}$ and centroid $\hat{\vec{c}}_{i}$, and generates $\tilde{\vec{\nu}}_i$ and $\tilde{\vec{c}}_j$, respectively. It should be noted that this operation does not produce any degradations in terms of the solution of the LP optimization problem. More specifically, $\hat{\vec{\nu}}_{i}$ can be calculated as
\begin{eqnarray*}
	\tilde{\vec{\nu}}_{i} 
	& = & \text{shrink}(\hat{\vec{\nu}}_{i})
	\ =\ (\hat{\vec{\nu}}_{i})_{| \mathcal{S}_{\rm samp}} \ \ \in \mathbb{R}^{|\mathcal{S}_{\rm samp}|}, 
\end{eqnarray*}
where $\mathcal{S}_{\rm samp} = {\rm supp}(\tilde{\vec{\nu}}_{i})$. Similarly to this, we calculate 
\begin{eqnarray*}
	\tilde{\vec{c}}_{i} & = & \text{shrink}(\hat{\vec{c}}_{i}) \ =\ (\hat{\vec{c}}_{i})_{| \mathcal{S}_{\rm cent}} \ \ \in \mathbb{R}^{|\mathcal{S}_{\rm cent}|}, 
\end{eqnarray*}
where $\mathcal{S}_{\rm cent} = {\rm supp}(\tilde{\vec{c}}_{i})$.

Accordingly, we must also shrink the ground cost matrix $\mat{C}_{\vec{\nu c}}$ based on the change of the size of $\tilde{\vec{c}}_{i}$ and $\tilde{\vec{\nu}}_{i}$. For this purpose, we newly introduce the {\it matrix shrinking} operator $\text{Shrink}(\cdot)$, which removes the row and the column vectors from $\mat{C}_{\vec{\nu c}}$. The removal is performed against $(\mathcal{S}_{\rm cent})^c$ and $(\mathcal{S}_{\rm samp})^c$. Consequently, $\mat{C}_{\vec{\nu c}}$ is compressed into $\tilde{\mat{C}}$ using $\text{Shrink}(\cdot)$ as
\begin{eqnarray*}
	\tilde{\mat{C}} = \text{Shrink}(\mat{C}_{\vec{\nu c}})
	 = \mat{C}_{{\rm supp}(\tilde{\vec{\nu}}_{i}),{\rm supp}(\tilde{\vec{c}}_{i})} \ \in \mathbb{R}^{ |\mathcal{S}_{\rm samp}| \times |\mathcal{S}_{\rm cent}|}.
\end{eqnarray*}

It should also be noted that the size of $\tilde{\mat{C}}$ is $ |\mathcal{S}_{\rm samp}| \times |\mathcal{S}_{\rm cent}|$. Those sizes change according to the control parameter of the sparse ratio $\gamma(t)$, which is described in the next subsection.

\subsection{Control parameter of sparse ratio $\gamma(t)$}
\label{sec:gamma}

As described in {Section \ref{Sec:SSP}}, the control parameter of the sparsity is denoted as $\gamma(t)$, where $t$ represents the iteration number $t$. We also propose three $\gamma(t)$ control algorithms, which are denoted respectively as `FIX' (fixed), `DEC' (decrease), and `INC'(increase). They are mathematically formulated as
\begin{equation*}
  \gamma(t):=\left\{
	\begin{array}{lrl}
    \displaystyle{\gamma_{\rm min}}&\quad&  \text{({FIX})} \\
    \displaystyle{1-\frac{(1 - \gamma_{\rm min})}{T_{\rm max}}t }&& \text{({DEC})} \\
    \displaystyle{\gamma_{\rm min} + \frac{(1 - \gamma_{\rm min})}{T_{\rm max}}t} && \text{({INC})},
  \end{array} 
  \right.
\end{equation*}
where $\gamma_{\rm min} \in \mathbb{R}$ is the minimum value, and where $T_{\rm max} \in \mathbb{N}$ is the maximum number of the iterations. The overall algorithm of the proposed SSPW $k$-means is presented in {\bf Algorithm \ref{Alg:SSPWkmeans}}.

\begin{landscape}
\begin{table*}[htbp]
\caption{Averaged clustering performance results on USPS dataset dataset (10 runs). The best result in each $\gamma_{\rm min}$ is shown in bold. The best in all settings is shown in bold with underline.}

\begin{center}
{\small
\label{tabl:USPS_dataset}
\begin{tabular}{c|c|c|c||c|c|c|c|c|c|c|c|c|c|c|c|c|c|c|c}
\hline
\!\!\!method\!\!\! & \!\!\!shrink\!\!\! & \multicolumn{2}{|c||}{\!\!\!projection\!\!\!} &  \multicolumn{4}{|c|}{Purity} & \multicolumn{4}{|c|}{NMI} &  \multicolumn{4}{|c|}{Accuracy} & \multicolumn{4}{|c}{Time } \\
\cline{3-4}
 &\!\!\!oper.\!\!\! & $\tilde{\vec{\nu}}_i$ & $\tilde{\vec{c}}_j$ & \multicolumn{4}{|c|}{[$\times 10^{2}$]} & \multicolumn{4}{|c|}{[$\times 10^{2}$]} & \multicolumn{4}{|c|}{[$\times 10^{2}$]}& \multicolumn{4}{|c}{[$\times 10^{2}$sec]}  \\
 \hline
\hline
\!\!\!\!\!\!\!\!{$k$-means}\!\!\!\!\!\!\!\! & & & & \multicolumn{4}{|c|}{{65.5}} & \multicolumn{4}{|c|}{{65.8}} & \multicolumn{4}{|c|}{{64.1}}& \multicolumn{4}{|c}{{$2.13 \times 10^{-4}$}}\\
\cline{1-4}
\hline
\hline
\!\!\!\!\!\!\!\!baseline\!\!\!\!\!\!\!\! & & & & \multicolumn{4}{|c|}{66.6} & \multicolumn{4}{|c|}{65.0} & \multicolumn{4}{|c|}{65.5}& \multicolumn{4}{|c}{8.29}\\
\cline{1-4}
\cline{17-20}
shrink & $\checkmark$ && & \multicolumn{4}{|c|}{} & \multicolumn{4}{|c|}{} & \multicolumn{4}{|c|}{}& \multicolumn{4}{|c}{7.02} \\
\hline
\hline
 --& --  &\multicolumn{2}{|c||}{$\gamma_{\rm min}$} & 0.5 & 0.6 & 0.7 & 0.8 & 0.5 & 0.6 & 0.7 & 0.8& 0.5 & 0.6 & 0.7 & 0.8& 0.5 & 0.6 & 0.7 & 0.8 \\
\hline
\hline
 & $\checkmark$& &$\checkmark$&
67.8 &67.0&{\bf 66.9}&66.8&
66.2&65.5&{\bf 65.9}&65.2&
66.4&65.8&{\bf 65.8}&65.8&
3.11&3.25&3.99&5.04
\\
\cline{2-20}
fix & $\checkmark$& $\checkmark$& &
66.0&{\bf 67.2}&66.2&66.4&
64.6&65.4&65.0&64.9&
65.0&{\bf 66.1}&65.1&65.3&
1.87&2.98&3.91&4.60
\\
\cline{2-20}
 & $\checkmark$& $\checkmark$& $\checkmark$ &
\underline{\bf 68.1} &{\bf 67.2}&66.3&66.6&
\underline{\bf 66.3}&{\bf 65.6}&65.0&65.0&
\underline{\bf 67.0}&65.9&65.2&65.5&
\underline{\bf 0.98}&{\bf 1.68}&{\bf 2.48}&{\bf 3.46}
 \\
\hline
 & $\checkmark$& &$\checkmark$  &
66.8&66.8&66.8&66.8&
65.4&65.4&65.3&{\bf 65.3}&
65.7&65.7&65.7&65.7&
4.12&4.45&5.65&6.29
 \\
\cline{2-20}
dec & $\checkmark$& $\checkmark$ & &
67.0&67.0&66.8&66.6&
65.3&65.3&65.3&65.0&
65.8&65.8&65.7&65.5&
5.22&5.82&6.13&7.37
\\
\cline{2-20}
 & $\checkmark$& $\checkmark$ & $\checkmark$&
66.8&66.8&66.8&66.8&
65.4&65.4&65.4&{\bf 65.3}&
65.7&65.7&65.7&65.7&
3.51&3.91&4.89&5.62
 \\
\hline
 & $\checkmark$& & $\checkmark$&
66.3&65.7&{\bf 66.9}&{\bf 66.9}&
65.0&64.8&65.4&65.2&
65.1&64.6&65.7&{\bf 65.9}&
9.74&\!\!10.47&\!\!11.29&\!\!12.04 
 \\
\cline{2-20}
inc & $\checkmark$& $\checkmark$ & &
66.8&66.4&66.3&66.2&
65.3&65.0&64.9&65.0&
65.6&65.3&65.2&65.1&
7.11&6.64&6.97&6.63
\\
\cline{2-20}
 & $\checkmark$& $\checkmark$ & $\checkmark$&
66.8&65.8&66.3&66.5&
65.4&64.8&65.2&64.9&
65.5&64.7&65.2&65.5&
6.06&7.05&7.48&9.71
 \\
\hline
\end{tabular}
}
\end{center}
\end{table*}
\end{landscape}

\begin{figure*}[t]
\begin{center}

\begin{minipage}[b]{.48\linewidth}
 \centering
\includegraphics[width=1\linewidth]{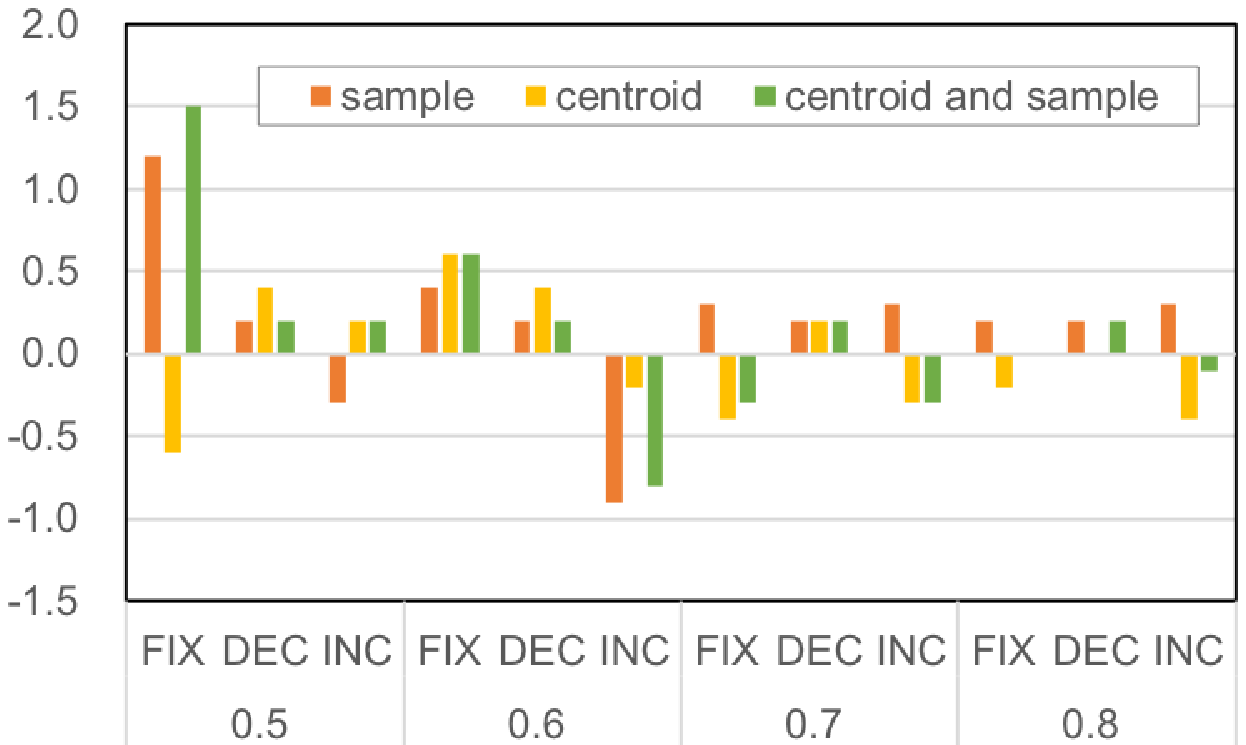}

 {\small (a) Purity}\medskip
\end{minipage}
\begin{minipage}[b]{.48\linewidth}
 \centering
\includegraphics[width=1\linewidth]{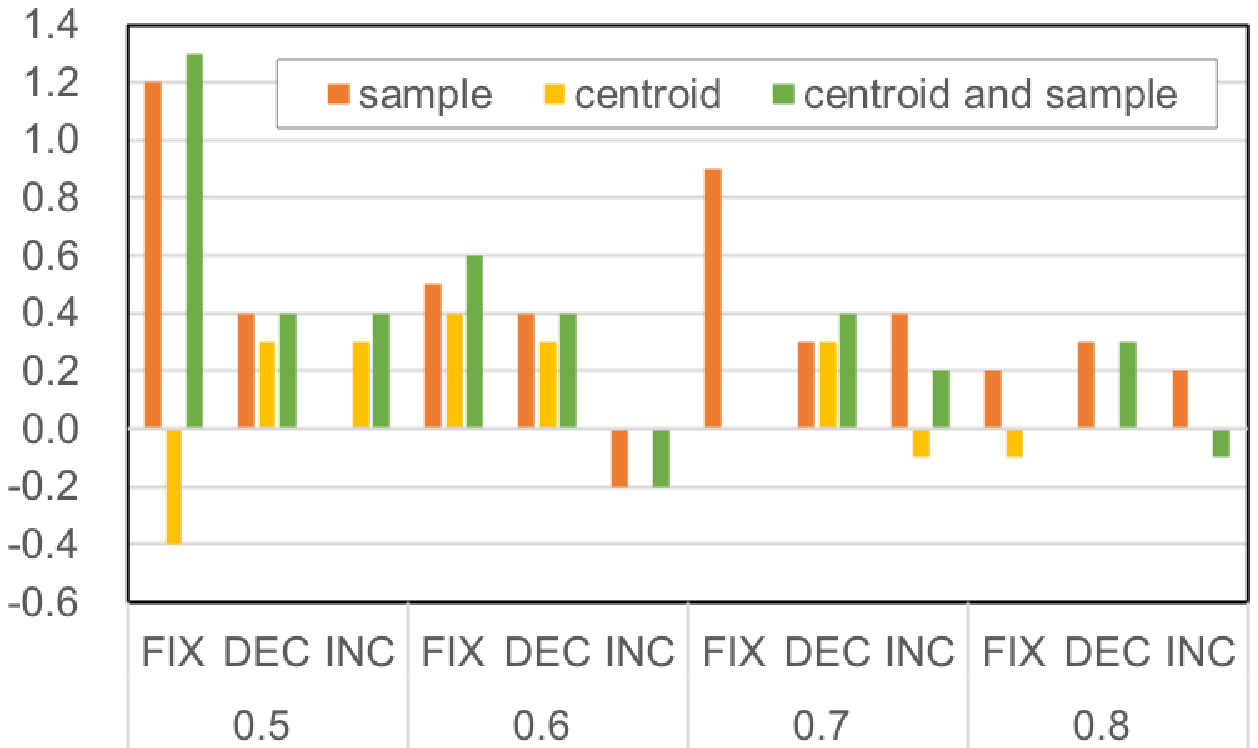}

{\small (b) NMI}\medskip
\end{minipage}

\begin{minipage}[b]{.48\linewidth}
 \centering
\includegraphics[width=1\linewidth]{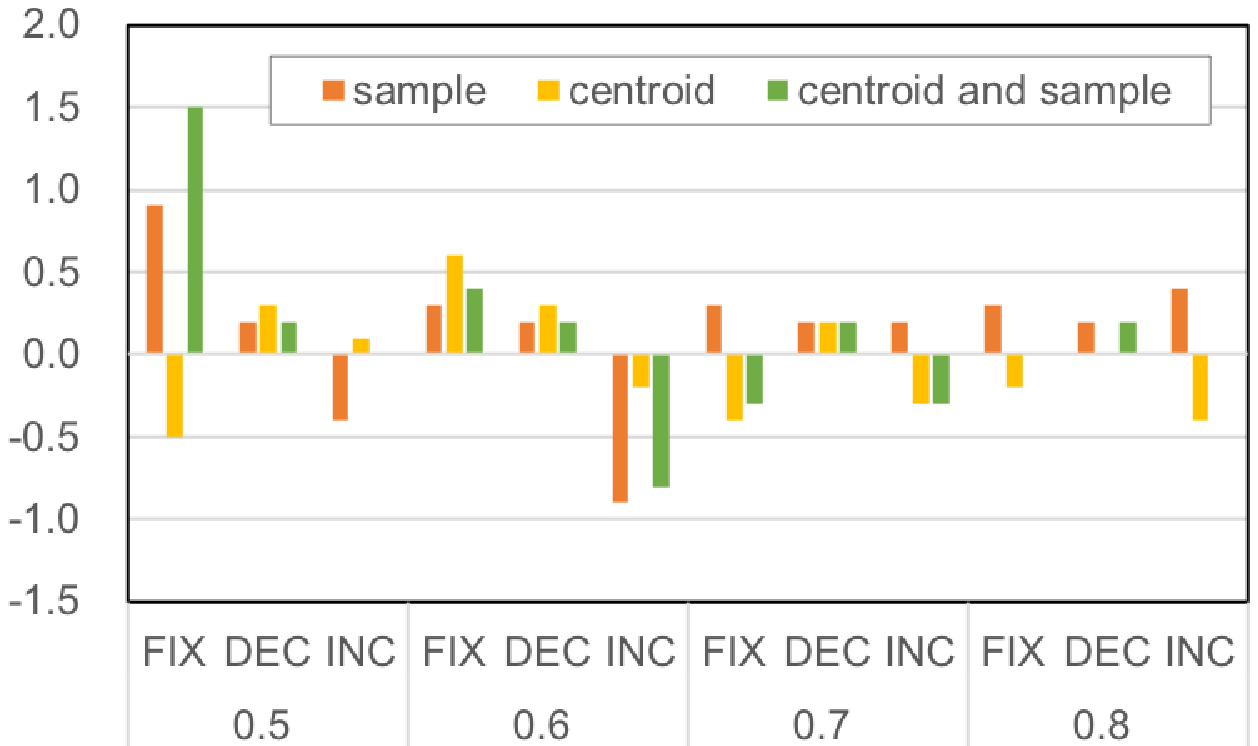}

 {\small (c) Accuracy}\medskip
\end{minipage}
\begin{minipage}[b]{.48\linewidth}
 \centering
\includegraphics[width=1\linewidth]{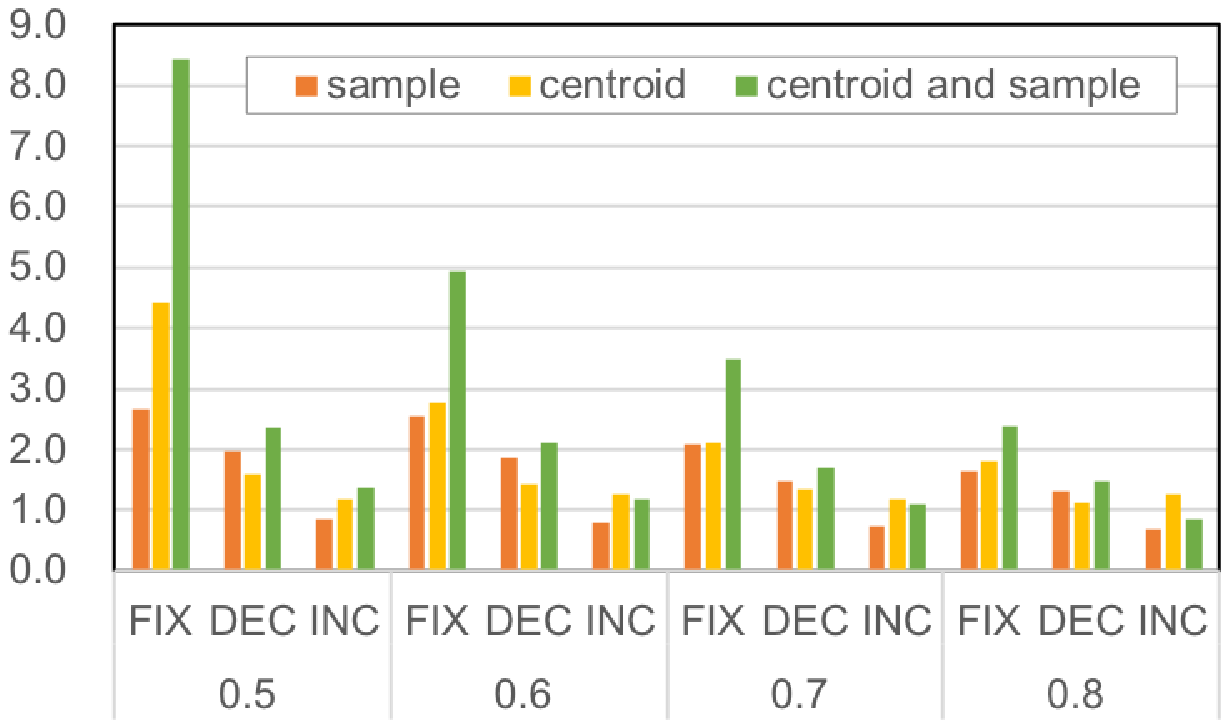}

{\small (d) Computation time}\medskip
\end{minipage}
\end{center}

\caption{Performance results of 2-D histogram data on the USPS dataset.}
\label{fig:USPS_clustering_results}
\end{figure*}

\begin{figure*}[t]
\begin{center}
\begin{minipage}[b]{.48\linewidth}
 \centering
\includegraphics[width=1\linewidth]{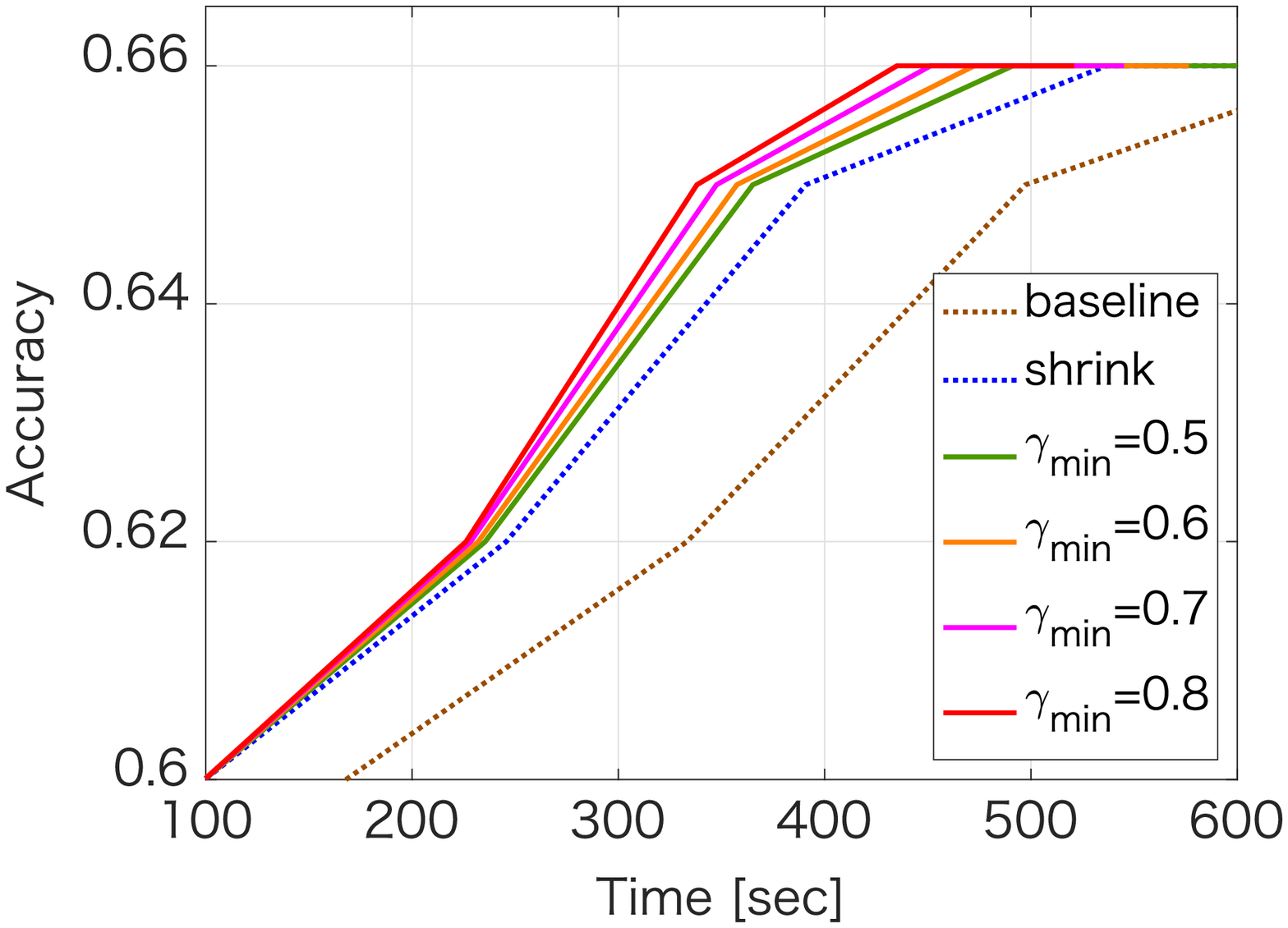}

{\small (a) projection of sample}\medskip
\end{minipage}
\begin{minipage}[b]{.48\linewidth}
 \centering
\includegraphics[width=1\linewidth]{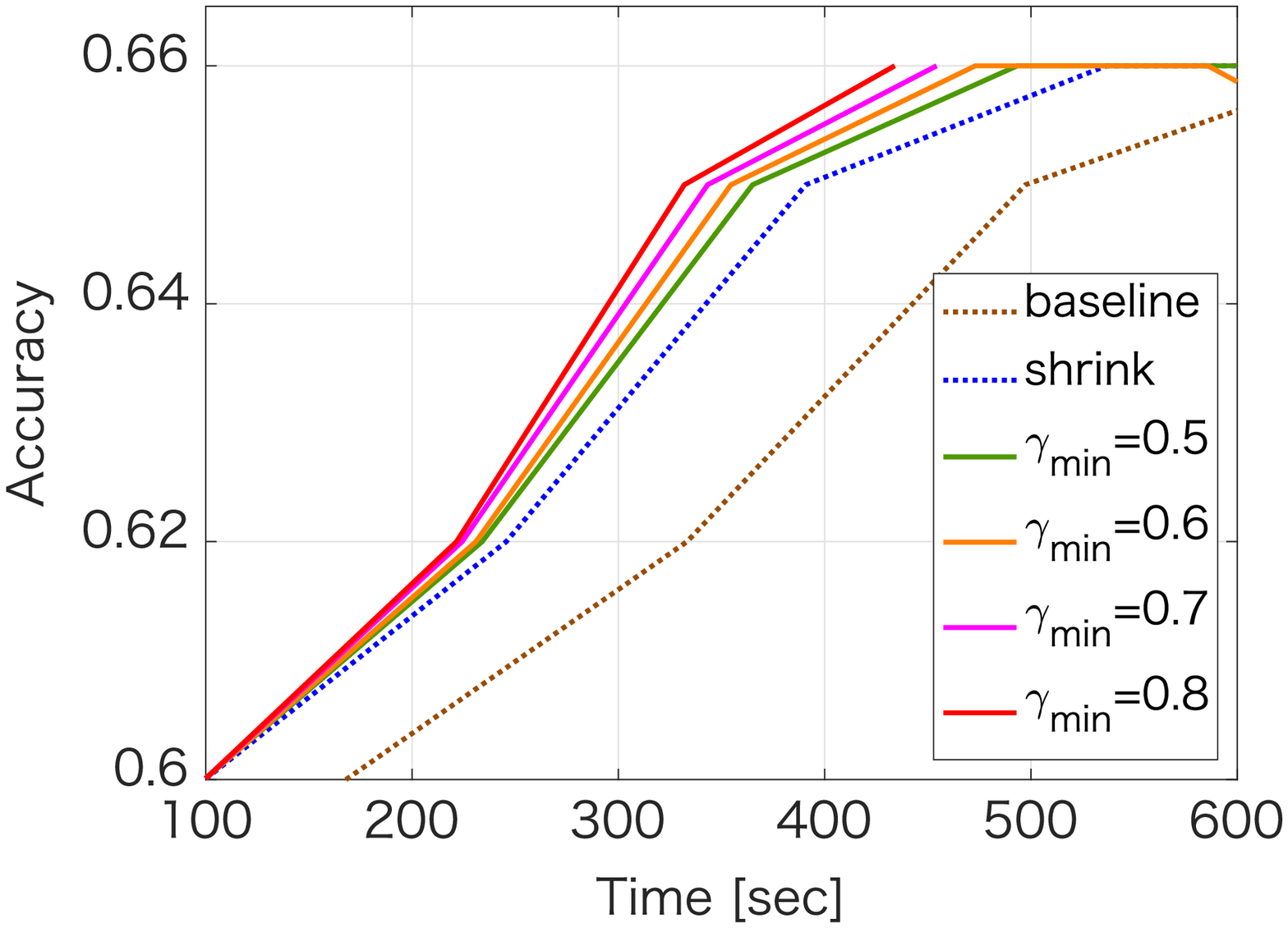}

 {\small (b) projection of centroid}\medskip
\end{minipage}

\begin{minipage}[b]{.48\linewidth}
 \centering
\includegraphics[width=1\linewidth]{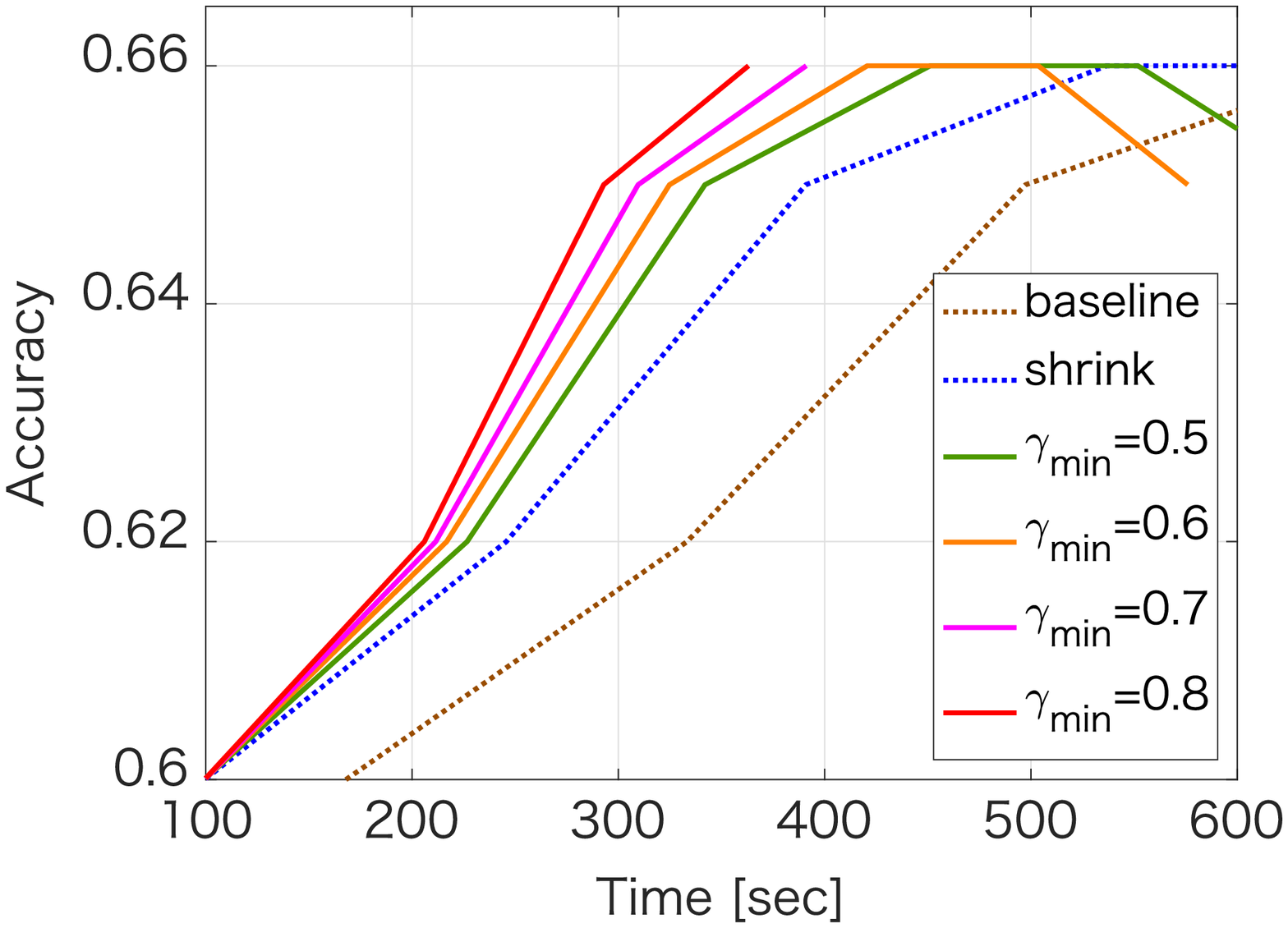}

{\small (c) projection of centroid and sample}\medskip
\end{minipage}
\end{center}

\caption{Convergence performance results for different projection data with different $\gamma_{\rm min}=\{0.5, 0.6, 0.7, 0.8\}$.}
\label{fig:USPS_diff_gamma}
\end{figure*}

\section{Numerical evaluation}
We conducted numerical experiments with respect to computational efficiency and clustering quality on real-world datasets to demonstrate the effectiveness of the proposed SSPW $k$-means. Regarding the Wasserstein barycenter algorithm, we use that proposed in \cite{Benamou_2015_SIAMJSC}. The initial centroid is calculated using the Euclidean $k$-means with {\sf litekmeans}\footnote{\url{http://www.cad.zju.edu.cn/home/dengcai/Data/code/litekmeans.m}.}. {\sf linprog} of Mosek \cite{MOSEK_2000}\footnote{\url{https://www.mosek.com/}} is used to solve the LP problem. Throughout the following experiment, the standard Wasserstein $k$-means described in {Section \ref{sec:Wkmeans}} is referred as {\it baseline} method. {The Euclidean $k$-means algorithm is also compared to evaluate the clustering quality metrics although its computation is much lower than the Wasserstein one.} We set $T_{\rm max}=10$. {The experiments were performed on a four quad-core Intel Core i5 computer at 3.6 GHz, with 64 GB DDR2 2667 MHz RAM. All the codes are implemented in MATLAB.} 

\subsection{1-D histogram evaluation}
We first conducted a preliminary experiment using the COIL-100 object image dataset\footnote{http://www1.cs.columbia.edu/CAVE/software/softlib/coil-100.php} \cite{COIL-100}, which includes $7,200$ images of $100$ objects. From this dataset, we randomly select $10$ classes and $10$ images per class. We first convert the pixel information into intensity with the range of $0$ to $255$. Then we generate a one-dimensional histogram of which the bin size is $255$ by removing the intensity of zero. We set $\gamma_{\rm min}=\{0.7, 0.8\}$.

The averaged clustering performances on Purity, NMI and Accuracy over $5$ runs are summarized in TABLE \ref{tabl:COIL100_dataset}. For ease of comparison, the difference values against the baseline method are presented in {Fig. \ref{fig:COIL100_clustering_results}}, of which panels (a), (b) and (c) respectively show results of Purity, NMI, and Accuracy. Here, {\it positive} values represent improvements against the baseline method. Additionally, {Fig. \ref{fig:COIL100_clustering_results}} (d) presents the speed-up ratio of the computation time against that of the baseline method, where the value more than than $1.0$ indicate faster computations than that of the baseline. The figures specifically summarize the results in terms of the combinations of different $\gamma(t)$ algorithms and different $\gamma_{\rm min}$. They also show the differences among the different projection types, i.e., the projection of the centroid, data sample, and both the centroid and data sample. From {Fig. \ref{fig:COIL100_clustering_results}}, one can find that the case with the centroid projection stably outperforming other cases. Especially, the INC algorithm with the centroid projection stably outperforms others. However, it requires a greater number of computations. Moreover, the effectiveness in terms of the computation complexity reduction is lower. On the other hand, the FIX algorithm with $\gamma_{\rm min}=0.7$ engenders comparable improvements while keeping the computation time much lower. 

\subsection{2-D histogram evaluation}
\label{Sec:2D_histogram_evaluations}
In this experiment, comprehensive evaluations have been conducted. We investigate a wider range of $\gamma_{\rm min}$ than that of the earlier experiment, i.e., $\gamma_{\rm min}=\{0.5, 0.6, 0.7, 0.8\}$. For this purpose, we used the USPS handwritten dataset\footnote{\url{http://www.gaussianprocess.org/gpml/data/}}, which includes completely $9298$ handwritten single digits between $0$ and $9$, each of which consists of $16 \times 16$ pixel image. Pixel values are normalized to be in the range of $[-1, 1]$. From this dataset, we randomly select $10$ images per class. We first convert the pixel information into intensity with the range of $0$ to $255$. Then we generate a {\it two-dimensional} histogram for which the sum of intensities of all pixels is equal to one.

The averaged clustering qualities over $10$ runs are presented in {TABLE \ref{tabl:USPS_dataset}}. As shown there, the FIX algorithm outperforms others in terms of both the clustering quality and the computation time. Similarly to the earlier experiment, for ease of comparison, the difference values against the baseline method are also shown in {Fig. \ref{fig:USPS_clustering_results}}. It is surprising that the proposed algorithm maintains the clustering quality across all the metrics as higher than the baseline method does, even while reducing the computation time. Addressing individual algorithms, although the INC algorithm exhibits degradations in some cases, the DEC algorithm outperforms the baseline method in all settings. Furthermore, as unexpected, the performances of the algorithms with lower $\gamma_{\rm min}$, i.e., $\gamma_{\rm min}=\{0.5, 0.6\}$, are comparable with or better than those with $\gamma_{\rm min}=\{0.7, 0.8\}$.

Second, the computation time under the same settings is presented in {Fig. \ref{fig:USPS_clustering_results}}(d). From this result, it is apparent that the computation time is approximately proportional to $\gamma_{\rm min}$ in each $\gamma(t)$ control algorithm. In other words, the case with $\gamma_{\rm min}=0.5$ requires the lowest computation time among all. Moreover, with regard to the different projection types, the case in which the sparse simplex projection is performed onto both the centroid and data sample requires the shortest computation time, as expected.

\subsection{Convergence performance}
\begin{figure}[t]
\begin{center}
\includegraphics[width=0.6\linewidth]{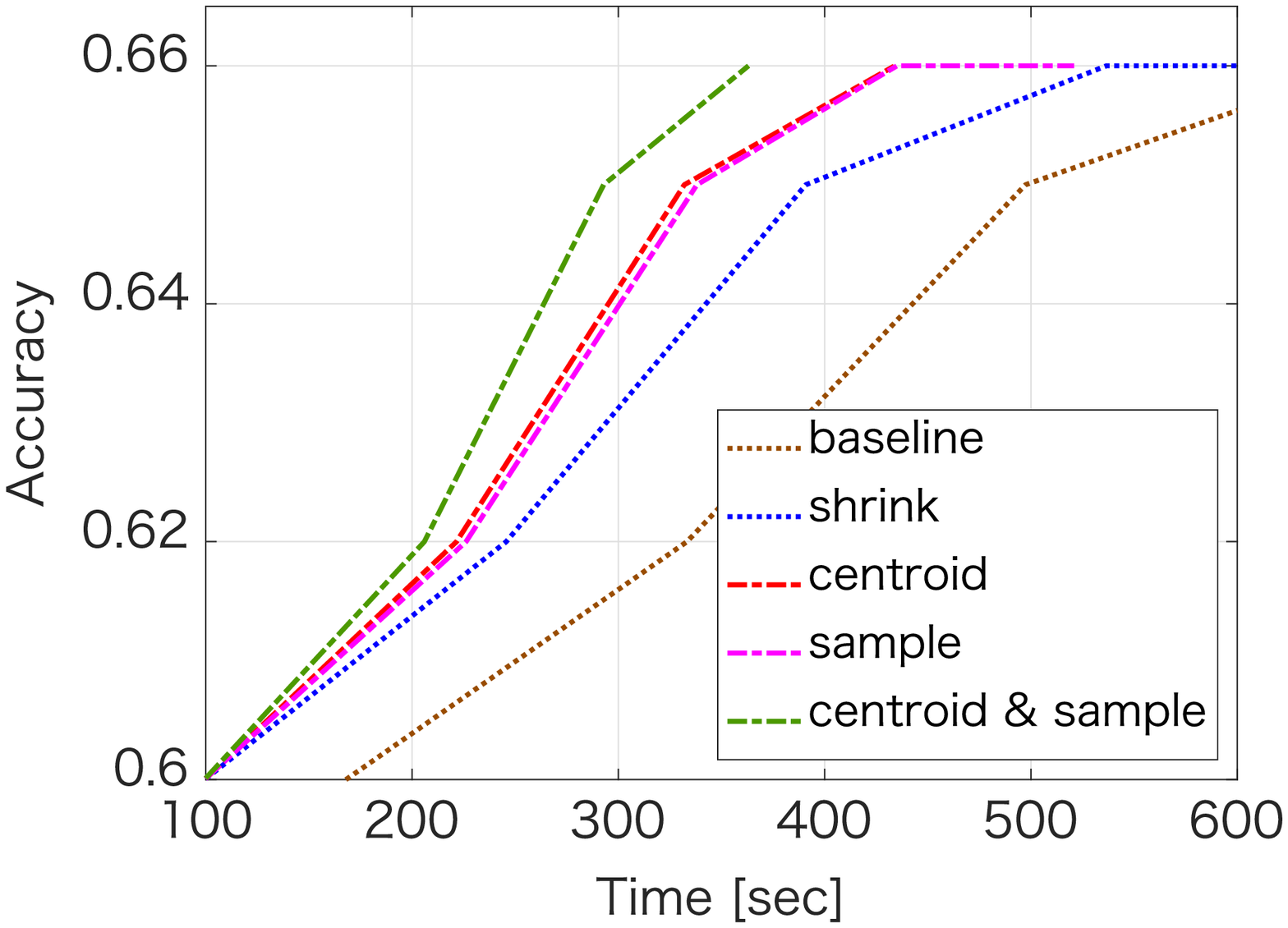}
\caption{\!Convergence with different projection data using $\gamma_{\rm min}\!=\!0.5$.\!}
\label{fig:USPS_diff_proj_type}

\includegraphics[width=0.6\linewidth]{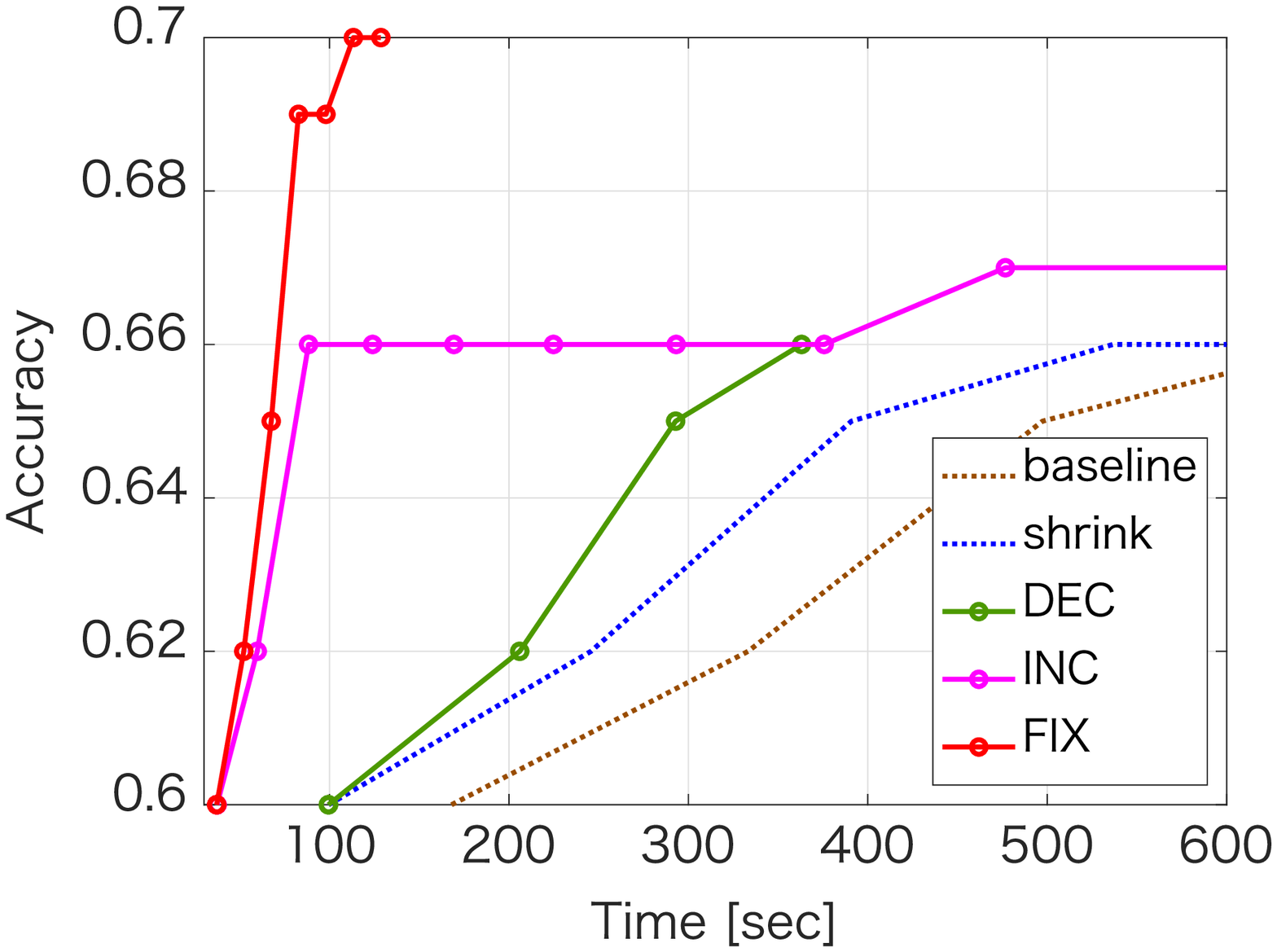}
\caption{Convergence  comparisons of different algorithms of $\gamma(t)$.}
\label{fig:USPS_diff_alg_results}
\end{center}
\end{figure}

This subsection investigates the convergence performances of the proposed algorithm. The same settings as those of earlier experiments are used. We first address the convergence speed in terms of the computing time with respect to the different projection types as well as different $\gamma_{\rm min}$. For this purpose, we examine the DEC algorithm. However, the two remaining algorithms behave very similar.  {Fig. \ref{fig:USPS_diff_gamma}} shows convergence of Accuracy in terms of computation time at the first instance of $10$ runs in the earlier experiment. {Fig. \ref{fig:USPS_diff_gamma}}(a)--{Fig. \ref{fig:USPS_diff_gamma}}(c) respectively show the results obtained when the projection is performed onto the data sample, centroid, and both the centroid and data sample. Regarding the different $\gamma_{\rm min}$, the case with the smaller $\gamma_{\rm min}$ indicates faster convergence. Apart from that, the convergence behaviors among three projection types are similar. However, the case with the which projection of both the centroid and data sample yields the fastest results among all. This is also demonstrated explicitly in {Fig. \ref{fig:USPS_diff_proj_type}} under $\gamma_{\rm min}=0.5$.

We also address performance differences in terms of the different algorithms of the $\gamma(t)$ control parameters. {Fig. \ref{fig:USPS_diff_alg_results}} presents the convergence speeds obtained when projecting both the centroid and data sample with $\gamma_{\rm min}=0.5$, where the plots explicitly show individual iteration. As might be readily apparent, the FIX control algorithm and the INC control algorithm require much less time than the others because these two algorithms can process much smaller centroids and data samples than others at the beginning of the iterations. However, the computation time of the INC control algorithm increases gradually as the iterations proceed, whereas that of the DEC control algorithm decreases gradually. Overall, the FIX control algorithm outperforms the other algorithms.

\subsection{Performance comparison on different ratios}
Finally, we compared performances on different sparsity ratios $\{0.2, 0.3, 0.4, 0.5,0.6, 0.7,0.8,0.9\}$ of $\gamma_{\rm min}$ under the same configurations of the earlier experiment. We specifically address the FIX algorithm of $\gamma(t)$ and the case where the projection is performed onto both the centroid and sample. Regarding the time comparison, we use the same speed-up metric against the baseline method as the earlier experiment. Therefore, the range of the values is more than $0.0$, and the value more than $1.0$ means a speed-up against the baseline method. For this experiment, we have re-conducted the same simulation, thereby the obtained value are slightly different from those in the earlier experiment due to its randomness. 

\begin{figure}[htbp]
\begin{center}
\includegraphics[width=0.8\linewidth]{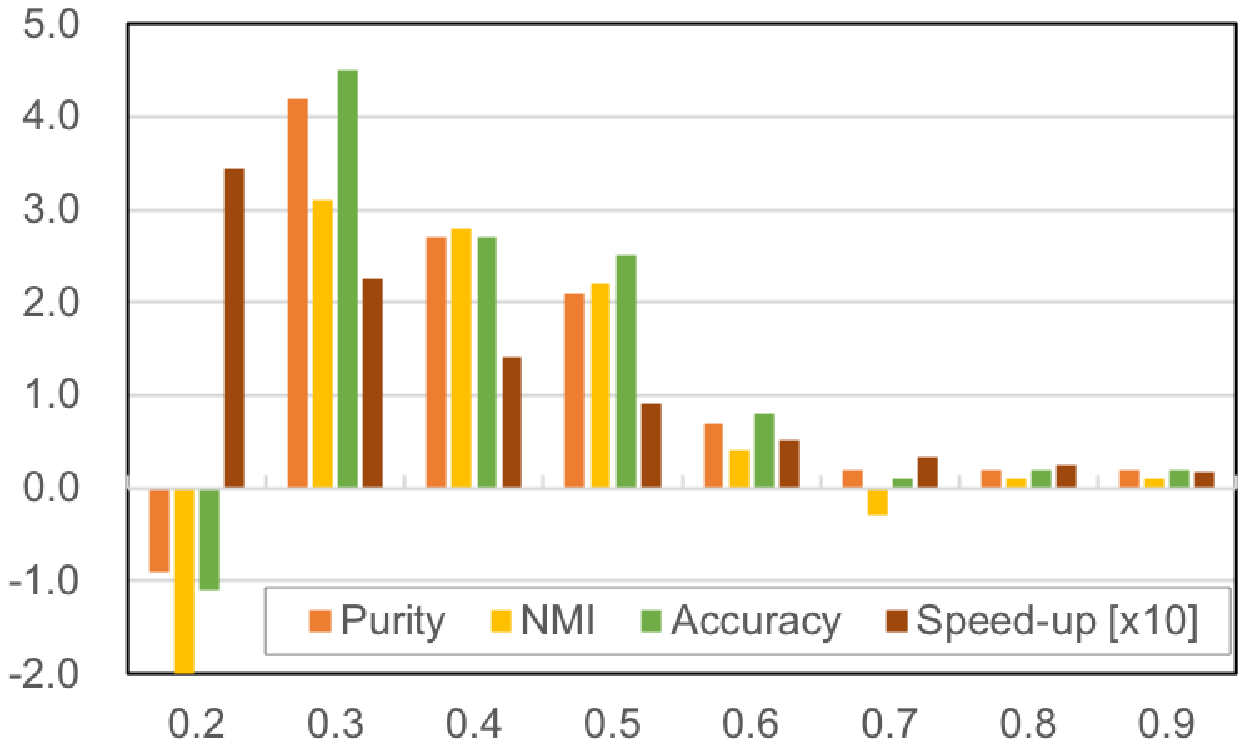}
\caption{Performance comparison on different ratios on the USPS dataset.}
\label{fig:USPS_Ratio_comparison_results}

\end{center}
\end{figure}

The results are summarized in {TABLE \ref{tabl:USPS_ratio_comparison_dataset}}. {Fig. \ref{fig:USPS_Ratio_comparison_results}} also demonstrates the results for ease of comparison. From both of the results, the cases with 
$\gamma_{\rm min}=\{0.3, 0.4\}$ yield higher performances than other cases with respect to the clustering quality as well as the computation efficiency. Surprisingly, we obtained around $4$ points higher values in the clustering quality metrics with more than $20$-fold speed-up. On the other hand, $\gamma_{\rm min}=0.2$ results in the worst performance. Therefore, the optimization of $\gamma_{\rm min}$ to give best performances is one of the important topics in the future research.

\begin{table}[htbp]
\caption{Averaged clustering performance results on USPS dataset dataset (10 runs). The best result in each $\gamma_{\rm min}$ is shown in bold.}

\begin{center}
\label{tabl:USPS_ratio_comparison_dataset}
\begin{tabular}{c|c||c|c|c|c}
\hline
method & $\gamma_{\rm min}$ & Purity & NMI & Accuracy & Time\\
 & & [$\times 10^{2}$] & [$\times 10^{2}$] & [$\times 10^{2}$] & [$\times 10^{2}$sec]\\
\hline
\hline
baseline & 0.00&69.7&69.6&66.9&7.98\\\cline{1-1}\cline{6-6}
shrink & &&&&6.77\\\hline\hline
&0.90&69.9&69.7&67.1&4.68\\\cline{2-6}
FIX&0.80&69.9&69.7&67.1&3.14\\\cline{2-6}
&0.70&69.9&69.3&67.0&2.34\\\cline{2-6}
 with & 0.60&70.4&70.0&67.7&1.54\\\cline{2-6}
shrink \&& 0.50&71.8&71.8&69.4&0.88\\\cline{2-6}
projection&0.40&72.4&72.4&69.6&0.56\\\cline{2-6}
of $\tilde{\vec{\nu}}_i$\&$\tilde{\vec{c}}_j$&0.30&{\bf 73.9}&{\bf 72.7}&{\bf 71.4}&0.35\\\cline{2-6}
 &0.20&68.8&67.6&65.8&0.23\\
\hline
\end{tabular}
\end{center}
\end{table}

\subsection{Discussions}
From the exhaustive experiments above, it can be seen that the proposed proposed SSPW $k$-means algorithm dynamically reduce the computational complexity by removing lower-valued histogram elements and harnessing the sparse simplex projection while maintaining degradation of the clustering quality lower. Although the performance of each setting of the proposed algorithm depends on the dataset, we conclude that the FIX algorithm with the projections of both the centroid and sample under lower sparsity ratios $\gamma_{min}$ is the best option because it gives stably better classification qualities with the fastest computing times among all the settings. In fact, this algorithm gives comparable performances to the best one in TABLE \ref{tabl:COIL100_dataset}, and outperforms the others in TABLE \ref{tabl:USPS_dataset}. The algorithm also yiedls around $4$ points higher values in the clustering quality metrics with more than $20$-fold speed-up as seen in {TABLE \ref{tabl:USPS_ratio_comparison_dataset}}.

\section{Conclusions}
This paper proposed a faster Wasserstein $k$-means algorithm by reducing the computational complexity of Wasserstein distance, exploiting sparse simplex projection operations. Numerical evaluations demonstrated the effectiveness of the proposed SSPW $k$-means algorithm. As for a future avenue, we would the proposed approach into a stochastic setting \cite{Kasai_JMLR_2018}.

\bibliographystyle{unsrt}
\bibliography{optimal_transport,dataset,clustering,nmf}

\end{document}